\documentclass[journal]{IEEEtran}

\usepackage[pdftex]{graphicx}

\ifCLASSINFOpdf
\else
\fi

\usepackage{array}
\newcolumntype{L}[1]{>{\raggedright\let\newline\\\arraybackslash\hspace{0pt}}m{#1}}
\newcolumntype{C}[1]{>{\centering\let\newline\\\arraybackslash\hspace{0pt}}m{#1}}
\newcolumntype{R}[1]{>{\raggedleft\let\newline\\\arraybackslash\hspace{0pt}}m{#1}}

\begin{document}
%

\title{Efficient CNN Implementation for Eye-Gaze Estimation on Low-Power/Low-Quality Consumer Imaging Systems}
%
%
%

\author{Joseph~Lemley,~\IEEEmembership{Student Member,~IEEE,}
        Anuradha ~Kar,~\IEEEmembership{Student Member,~IEEE,}
        Alexandru ~Drimbarean,~\IEEEmembership{Member,~IEEE,}
        and~Peter~Corcoran,~\IEEEmembership{Fellow,~IEEE}
\thanks{Joseph Lemley, Anurada Kar, and Peter Corcoran are with the Department
of Electrical and Electronic Engineering, National University of Ireland Galway, Galway,
Ireland e-mail: J.lemley2@nuigalway.ie
Alexandru Drimbarean is with Xperi Corporation, Galway Ireland, 
 }
\thanks{}}

%
%

\markboth{}%
{Shell \MakeLowercase{\textit{et al.}}: Bare Demo of IEEEtran.cls for IEEE Journals}
%


\maketitle

\begin{abstract}

Accurate and efficient eye gaze estimation is important for emerging consumer electronic systems such as driver monitoring systems and novel user interfaces. Such systems are required to operate reliably in difficult, unconstrained environments with low power consumption and at minimal cost.  In this paper a new hardware friendly, convolutional neural network model with minimal computational requirements is introduced and assessed for efficient appearance-based gaze estimation. The model is tested and compared against existing appearance based CNN approaches, achieving better eye gaze accuracy with significantly fewer computational requirements. A brief updated literature review is also provided. 

\end{abstract}


%
\IEEEpeerreviewmaketitle

\section{Introduction}

The potential of eye gaze tracking and gaze-based human computer
interactions in modern consumer devices is currently an active topic for exploration. Eye gaze has been used to derive human behavioral cues, as an input modality and for achieving immersive user experiences in virtual and augmented reality systems. However, applications of gaze in consumer devices operating in real world conditions face tough challenges in terms of accuracy and reliability

\subsection{Gaze Tracking in Consumer Devices}

After decades of research on desktop-based gaze estimation techniques, the focus has recently shifted to building eye gaze applications for dynamic platforms such as driver monitoring systems \cite{1} and handheld devices \cite{2}. For an automobile driver, eye based cues such as levels of gaze variation, speed of eyelid movements and eye closure can be indicative of a driver's cognitive state. These can be useful inputs for intelligent vehicles to understand driver attentiveness levels, lane change intent, and vehicle control in the presence of obstacles to avoid accidents \cite{3}. 

Handheld devices like smartphones and tablets form unique platforms for gaze tracking applications wherein gaze may be used as an input modality for device control, activating safety features and novel UI designs \cite{4}. 
The most challenging aspect of these modern gaze applications includes operation under dynamic user conditions and unconstrained environments. Further requirements for implementing a consumer-grade gaze-tracking system include real-time high-accuracy operation, minimal or no calibration, and robustness to user head movements and varied lighting conditions. Therefore accurate and reliable gaze tracking typically demands high quality cameras and special equipment like narrow angle lenses, external illumination, and stereo setups \cite{5} for capturing eye region features with sufficient details. As a result, gaze estimation systems frequently become costly with complicated setups, which are unsuitable for generic and consumer applications. 

Therefore a major challenge of gaze based consumer electronics design involves maximizing system performance while reducing costs and system complexities.

\subsection{Deep Learning for Eye Gaze}

In this paper, we introduce a calibration-free method for appearance-based gaze estimation that is suitable for consumer applications and low cost hardware with real time requirements, using a Convolutional Neural Network (CNN). 

Convolutional Neural networks (CNNs) were popularized by Lecun et al. \cite{lecun1989backpropagation}, who used them successfully for handwritten digit classification. These networks are inspired by the organization of the visual cortex and allow spatial information to be more efficiently learned. Convolutional Neural Networks can be used on input with any number of dimensions, but due to their success in pictures, are most popularly implemented for 2D input plus color channels. Other popular types of CNN's include 1D CNNs, which are commonly used for time series, and 3D CNNs, which can be used for volumetric data or time series data where the third dimension represents either spatial frames or temporal frames \cite{lemley2017deep}.  Although CNNs have become ubiquitous for most computer vision tasks, they have yet to become popular for eye gaze estimation. 
\subsection{Contributions of this Work}

From the perspective of developing a deep learning model for gaze estimation, the task can either be considered as a regression task or a classification task. Although both are useful, regression provides the greatest predictive flexibility and thus this paper treats the eye gaze estimation task as a regression problem with the goal of finding a gaze angle ($\phi$,$\theta$) that corresponds with a low resolution eye image such as one taken from a distance with a simple RGB webcam mounted on a dashboard.  

In this paper, a hardware optimized network is implemented with demonstrated suitability for deployment on such consumer devices in terms of memory requirements and speed. This network achieves superior accuracy using a dual channel input technique when compared to other state-of-the-art CNN-based gaze tracking methods for unconstrained, low resolution eye tracking. 

\section{Related work}

In this section a review of conventional gaze tracking techniques, studies on using low resolution data, and the application of deep learning in gaze estimation are discussed. The development and usage of important databases for gaze research are also presented.

\subsection{Contemporary Methods for Eye Gaze Estimation}

Gaze-tracking algorithms can be broadly classified into two types: model-based methods and appearance-based methods. \cite{hansen2010eye} Appearance-based methods operate directly on the eye images.  

Examples of model-based methods include 2D and 3D models that use NIR illumination to create corneal reflections and track them with respect to the pupil center to estimate the gaze vector. These require polynomial or geometric approximations of the human eye to obtain the gaze direction or the point of gaze. Appearance-based methods use eye region images to extract content information such as local features, shape, and texture of eye regions to estimate gaze direction. Some key works in each of these classes of methods are summarized below.  

\subsubsection{Measures of Accuracy for Eye Gaze Tracking Methods}

Contemporary research on gaze tracking measures accuracy in a wide variety of ways \cite{Kar2017}. For example, commonly used measures include angular resolution in degrees \cite{coutinho2012augmenting}, gaze recognition rates in percentage \cite{chen2015eye}, and shifts in number of pixels or distance in cm/mm between gaze \cite{22} and target locations. Unfortunately, these 4 methods are not correlated and not inter-comparable. It is the view of the authors that angular resolution is most reliable as it describes the performance of an algorithm irrespective of other system variables like user distance from tracker, pixel size of screen etc. Therefore, in this work, the angular resolution in degrees is estimated and used as the metric of accuracy for the proposed algorithm. Results from this work are only directly compared to other papers which express their performance in angles. 

\subsubsection{2D Models}

2D models utilize polynomial transformation functions for mapping the gaze vector (vector between pupil center and corneal glint) to corresponding gaze coordinates on the screen. In \cite{6}, it is shown that calibration targets and components of the mapping function are significant in determining overall accuracy of a regression-based tracker.  Artificial Neural Network (ANN) based mapping methods are presented in \cite{7,8,11} . In \cite{11} a three layer ANN achieves better accuracy than regression-based approaches. Methods that are robust to head pose are presented in \cite{9} and \cite{10}.   Multiple geometrical transformation based mapping for handling variable user distances and head motion are used in \cite{9}. A highly accurate calibration-free algorithm  with tolerance for natural head movements is discussed in \cite{10} using a support vector machine (SVM). Another high resolution, head-pose-invariant tracking method that does not require geometric models is presented by \cite{12}. The typical accuracy of such models is between two and four degrees.

\subsubsection{3D Models}
3D model-based methods typically use a geometrical model of the human eye to estimate the center of the cornea, and the optical and visual axes of the eye.  Gaze coordinates are estimated as points of intersection of the visual axes with the scene. These methods achieve high accuracy (~1 degree) but require elaborate system setups and knowledge about geometric relations between system components like LEDs, monitors and cameras. \cite{13} presents a mathematical model to estimate the optical and visual axes of the users' eyes from the center of the pupil and glint, considering single and multiple cameras and light sources. Methods achieving high accuracy and head pose robustness are reported in \cite{5,14,15,16} which require multiple cameras in their setup. \cite{15} also uses a dynamic head compensation model for updating the mapping function to track gaze under natural head movement. Calibration-free gaze estimation techniques are proposed by \cite{17,18} in which cameras, light sources and a spherical model of the cornea are used.  Recent developments in 3D gaze tracking include usage of depth sensors along with RGB cameras such as proposed in \cite{19}. In this, 3D gaze coordinates can be tracked in real time with a Kinect device, which provides 3D coordinates of eye features while eye parameters like eyeball and pupil center are derived from user calibration. The Kinect sensor is used in \cite{20} for gaze estimation using free head motion along with the iris center localization method and geometric constraints-based eyeball center estimation. 

\subsubsection{Appearance-Based Methods}

Appearance based methods utilize cropped eye images of a subject gazing at known locations to generate gaze point coordinates. The eye images are then used as training data for various machine learning models. For example, in \cite{21}, coordinates of eye contours, iris size, location, and pupil positions are estimated using an Active Appearance Model (AAM) and then used as input to a support vector machine (SVM) for gaze estimation. In \cite{22}, texture features are obtained using Local-Binary-Pattern (LBP) and used with an SVM along with space coordinates of the eyes for head pose free gaze tracking.  A comparative evaluation of different classification methods, such as SVM, neural networks, and k-Nearest Neighbor (k-NN)s is presented in \cite{23} where Local Binary Patterns Histograms (LBPH) and Principle Component Analysis (PCA) are used to extract eye appearance features.  The use of Haar features is reported in \cite{24,25} for real time gaze tracking.  In \cite{27} a neural network with a skin colour model to detect face and eye regions is used. Head-pose tolerant tracking is achieved in \cite{26} using a neural network. Recently, appearance-based methods implemented using deep learning (DL) and convolutional neural network (CNN) approaches have gained momentum. These are described in detail in section C. 

\subsection{Gaze Estimation from Low Resolution Images}

To facilitate gaze tracking in everyday settings, the use of cheap, compact and easy-to-integrate webcams is commonly preferred. Unfortunately, webcams offer low resolution images (typically 640x480 pixels) resulting in very poor gaze estimation accuracy. Low resolution images have strong noise effects \cite{28}, and distortions in the eye region contours and eye features become indistinguishable under varying illumination levels, user distance, and movements. Therefore, several approaches have been developed to achieve high gaze accuracy from low quality images and are discussed in this subsection.

An early ANN approach was used to map gaze coordinates to low quality cropped eye images in \cite{26}. The ANN used back propagation and 50 output units each for X and Y coordinate. It was trained with 2000 image/gaze position pairs. A hybrid approach is adopted in \cite{29} in which the iris centres are determined first using circular a Hough transform, followed by refinement using a gradient-aware random sample consensus (RANSAC) algorithm and ellipse fitting. Eye corners are estimated using Gabor jets \cite{30} and tracked using optical flow with normalized cross-correlation. Finally, the point of gaze (POG) is estimated from the iris center and eye corners using regression. A similar method is proposed in \cite{28}, where the problem arising due to the small size of cropped eye regions from low resolution images is overcome using 2D bilinear interpolation for reconstructing the eye image to a larger size for accurate tracing of the corneal reflection vector.

In \cite{31}, multiple miniature low resolution cameras positioned around a head-mounted setup are used. The gaze-mapping function is learned from multiple cameras using a 512 unit ANN, and trained on a large dataset of eye images. In \cite{32}, the eyeball and its movement direction are detected using a deformable angular integral search (DAISMI) method followed by a Deformable template-based 2D gaze estimation (DTBGE) algorithm used as a noise filter. \cite{33} trains an appearance model using Singular Value Decomposition (SVD) and a set of eye region images expanded artificially by adding positioning errors. Then a third order tensor is estimated from the training images as the gaze direction and positioning vectors of these images. The SVD is trained and tested by extracting the gaze vector from the test images and comparing with that obtained from the training images.

\subsection{Eye Gaze Estimation Using CNN's}

Deep learning (DL) techniques have been successfully used in challenging conditions such as those with variable illumination, unconstrained backgrounds and free head motion. For example, \cite{36} describes a calibration-free real-time CNN-based framework for gaze classification. Two CNNs, for the left and right eyes, are then trained independently to classify the gaze in seven directions. In \cite{37}, CNN-based gaze tracker for Augmented and Virtual reality devices achieves foveated rendering and gaze-contingent focus. The deep learning model is built to be robust to variations like skin and eye colour, illumination  and occlusion. In \cite{38}, deep features are obtained from eye images using multi-scale convolutions and pooling for predicting gaze direction. This method uses minimized cross-entropy loss, coupled with Random Forest regression as a clustering algorithm. It classifies areas on a device screen according to gaze locations, and operates under natural illumination and head poses. \cite{41} describes a novel, appearance-based gaze estimation method in which a CNN utilizes the full face image as input with spatial weights on the feature maps to suppress or enhance information in different facial regions. It achieves high accuracy and robust performance under varied illumination and extreme head poses.  \cite{42} achieves free head pose, 3D gaze tracking using two separate head pose and eye movement models with two CNNs, connected via a “gaze transform layer. Finally, in \cite{39} a CNN is built to learn the mapping between 2D head angle, eye image and gaze angle (output) using a small Lenet-inspired CNN.  For testing, an extensive database is built with more than 200,000 images under variable illumination levels and eye appearances. This database (called MPII Gaze) is also used in this work and its further details are provided in the next section. 

Several of the CNN-based works are specifically targeted towards gaze tracking in consumer/handheld devices such as \cite{43, 44}. In \cite{43}, a CNN-based real time, calibration-free gaze estimation algorithm is presented. It is trained using a large and diverse dataset of eye images taken under variable lighting, head pose, and backgrounds captured from users through a smartphone app. Inputs to their CNN model include eye and face images.  The location of the faces in the images are obtained through a face grid, which is used to infer relative eye and head poses. \cite{44} presents a calibration-free method using Deep Belief Networks which classifiy gaze into a grid of nine gaze locations under various head-poses and viewing directions. In \cite{45}, a nine directional CNN-based gaze classifier is developed for a screen typing application, robust to false detections, blinks, and saccades (rapid, abrupt changes in fixation). 

An overview of selected deep learning methods for gaze estimation can be seen in table \ref{Selectedcnn}.
\begin{table*}[]
\centering
\caption{Selected CNN models for eye gaze estimation}
\label{Selectedcnn}
\begin{tabular}{|p{1cm}|p{3cm}|p{4.9cm}|p{1cm}|p{3.2cm}|p{2cm}|}
\hline
\textbf{Citation} & \textbf{Dataset used}  & \textbf{Network model used}   & \textbf{Image Resolution}  & \textbf{Accuracy}  & \textbf{Special features}  \\ \hline
\cite{37}                & CAVE and Own dataset (cropped images of the eye and their respective gaze pixel coordinates)                                        & LeNet (two convolutional layers and two pooling layers followed by a fully connected layer at the end). 21 classes for CAVE , 1829 classes for captured dataset. & 28x28                      & 6.7 degrees tested on CAVE as well as own collected dataset                                                    & Near eye tracking, operating under lighting changes and occlusions           \\ \hline
\cite{41}                & MPII Gaze, UT Multiview                                                                                                             & AlexNet. five convolutional layers, two fully connected layers. Additional linear regression layer on top of last fully connected layer.                               & 448 x 448  (full face) & 4.8 degrees on MPII Gaze, 6 degrees on UT dataset. Person-independent evaluation                                & Tolerates various illumination, gaze direction,                              \\ \hline
\cite{38}                & Own dataset with 107,681 images under different lighting, head movement, glasses.                                                   & Three convolutional layers with max- pooling layer, one single max-pooling layer, one hidden layer and one soft-max layer                                              & 40x70                     & 5-7 degrees within dataset evaluation, but training and test set have different gaze angles.                     & Works under natural light with free head motion                              \\ \hline
\cite{42}                & Own dataset 200-subjects different head poses, eyeball movements, lighting conditions, glasses, occlusions, reflections.           & AlexNet, BN-Inception network. Two CNNs to model head pose and eyeball motion. A gaze transform layer to aggregate them into gaze prediction                          & 62x62                     & 4.3 degrees cross subject evaluation                                                                           & Allows free head motion                                                      \\ \hline
\cite{39}                & MPII Gaze, Eyediap, UT Multiview                                                                                                    & LeNet architecture. A linear regression layer trained on top of fully connected layer. Multimodal CNN model to use eye image and head pose information                 & 60x36                      & $\sim$6 degrees Cross Dataset Evaluation                                                                       & Variable appearance, illumination, head pose                                 \\ \hline
\cite{43}                & Gaze Capture dataset & AlexNet + SVR  & 80x80  & 2.58cm (within dataset evaluation)  & Tolerant pose, appearance, and lighting \\ \hline
\cite{46}                & Own dataset. 56 groups of eye videos, 181440 eye images from 22 subjects                                                            & 3 convolutional layers followed by max-pooling layers, 2 fully connected and a soft-max layer for classification. 6 \& 54 classes                                      & 40x72               & 6.375 deg, 69.3\% Cross dataset evaluation                                                                     & ----                                                                         \\ \hline
\end{tabular}
\end{table*}

\begin{table*}[]
\centering
\caption{Publicly available datasets for gaze estimation}
\label{pubdata}
\begin{tabular}{|C{1.6cm}|C{.5cm}|C{2cm}|C{5.5cm}|C{1.3cm}|C{4cm}|}
\hline
\textbf{Name}             & \textbf{Persons} & \textbf{Items}                                                                                                & \textbf{Conditions}                                                                                                                                                                                                                                   & \textbf{Resolution}                          & \textbf{Purpose}                                                                                              \\ \hline
MPII Gaze \cite{39}        & 15                   & 213,659 images. & Software running on subjects' laptops ask participants to look at a random 20 on-screen positions and confirm                                                                                                                                                  & unknown                                     & Appearance-based gaze estimation in the wild.                                                                 \\ \hline
UT Multiview \cite{56}     & 50                   & 64000 eye images                                   & 160 gaze directions per person were acquired using 8 cameras (views)                                                                                                                                                                                         & SXGA resolution (1280x1024)                  & Training and test data for appearance-based gaze estimation methods.                                          \\ \hline
EYEDIAP \cite{57}          & 16                   & 94 sessions                                                                                                   & Diversity of participants, head poses, gaze targets and sensing conditions. Screen or 3D objects. Data collection with Kinect for RGB and depth video streams                                                                                                 & 640x480 at 30 fps                          & Training and evaluation of gaze estimation approaches with robustness to pose, person.                        \\ \hline
Gaze Capture \cite{43}      & 1474                 & 2445504 images                                                                                                & Data captured with iphone/ipad using app. Large variation in pose, appearance, lighting. Variation in relative distance and orientation of the mobile device                                                                                                  & unknown                                     & Training CNNs for high accuracy calibration-free eye tracking on handheld devices under variable conditions. \\ \hline
TabletGaze \cite{59}       & 51                   & 100000 images                                                                                                 & Video sequences recorded with tablet front-facing camera while subjects look at a dot on tablet screen. Unrestricted subject motion, each subject performed 4 body postures: standing, sitting, slouching, and lying.                                    & 1280x720                                   & Mobile gaze dataset for studying unconstrained mobile gaze estimation                                         \\ \hline
Weidenbacher \cite{61}     & 20                   & 2220 images                                                                                                   & Manual landmarks on  pupils, eye corners, nose tip, mouth corners. Horizontal head rotations (0° to 90° in steps of 10° ), vertical head poses 0°, 30°, and 60° azimuth (-20°,  +20°) elevation. For each head pose, nine different gaze conditions           & 1600x1200                                    & Evaluating computational methods for head pose and eye gaze estimation                                        \\ \hline
McMurrough \cite{62} & 20                   & 120 sessions                                                                                                  & Videos recorded eye motion as subjects look at, or follow a set of predefined points on a computer screen. Head position in 3D captured using a Vicon Motion Tracking System                                                                                  & 768x480 pixels at a frame rate of 29.97 Hz & To be used as a benchmark for Point of Gaze (PoG) detection algorithms                                        \\ \hline
OMEG \cite{63}             & 50                   & 40000 images                                                    & Eye images captured under multiple head poses, three fixed poses, 0 and ±30 degree , and a free pose style.                                                                                                                                          & 1280 x 1024                                  & Evaluating and comparing gaze tracking algorithms                                                             \\ \hline
HPEG \cite{65}             & 10                   & 20 videos                                                                                               & Subject faces camera frontally, free to move, background covers a big part of the image, with intense human action.                                                                                                                                           & 1280x960 pixels, 30 fps                      & Head pose and gaze estimation algorithm testing                                                               \\ \hline
\end{tabular}
\end{table*}

\subsection{Related Work Utilizing the MPII Gaze Dataset}

The MPII Gaze dataset\cite{39} is large and challenging, containing images collected under a wide range of realistic scenarios, such as varied illumination levels, eye appearances and head poses. Use of MPII gaze dataset for training and testing gaze estimation algorithms can be found in their own paper \cite{39}, as also in \cite{41,46, 48, 49}. \cite{39}, which introduces the MPII Gaze dataset, uses a multimodal CNN for gaze estimation and reports a cross dataset test accuracy of 6 degrees. \cite{41} uses full face (instead of eye only and multi-region) images with a CNN and achieves a person independent accuracy of 4.8 degrees on MPII Gaze while being robust to illumination variations and extreme head poses.  In \cite{46}, gaze location over a block of screen area is tracked using a CNN, and the cross-subject performance is tested with MPII Gaze and that authors' own dataset.  On  MPII Gaze, the classification accuracy is poor (75.6\%) compared to that on authors' dataset (92.5\%).   A Deep Regression Bayesian Network described in \cite{49} achieves an accuracy of 7.1 degrees when tested on this dataset.
In \cite{50,51,52}, the MPII Gaze dataset is used for comparing synthetic datasets and facial models. \cite{50} presents a method for synthesizing a large set of variable eye region images with a generative 3D eye region model. Then a gaze estimation method using the k-Nearest-Neighbour algorithm) is tested on the synthetic data and the MPII Gaze dataset to achieve an accuracy of 9.95 and 9.58 degrees respectively. Another method for synthetic, labelled photo-realistic eye region image creation is described in \cite{51} using head scan geometry. The generated dataset, along with MPII gaze, is used to test and compare the accuracy of a CNN based gaze estimation method.  In \cite{52}, a facial behaviour analysis tool is developed that is capable of tracking gaze vectors using a Conditional Local Neural Fields (CLNF) framework by detecting eye region features like eyelids, irises and pupils. The tool is tested on MPII Gaze to achieve 9.96 degrees of accuracy in gaze estimation. In \cite{53} a semi-supervised learning method is developed for improving the realism of simulated data and used to create refined training images for gaze estimation using CNNs. The CNNs are then tested on the MPII gaze dataset to achieve an error of 7.8 degrees.
Apart from the above, the MPII dataset has been used for training a CNN for gaze estimation coupled with gaze target discovery in \cite{54} and a cascaded-regressor-based eye center detector \cite{55}.  
 
\section{Image and video datasets for low resolution and unconstrained gaze estimation}

A survey of relevant publicly available gaze databases is summarized in Table \ref{pubdata}. In this survey, only the databases built for training and testing gaze estimation algorithms are listed, while other gaze databases, e.g. for studying saliency models, are not included as they are out of scope for this work.

\section{Methods}

The CNN-based gaze estimation methods in this work were evaluated on NVIDIA 1080 TI GPUs using python 2.7 and caffe 1.0 with accuracy and euclidean loss layers modified to calculate angle difference in radians. Person-exclusive, leave-one-out cross-validation was used in all experiments. 

In eye gaze tracking literature it is common to use the word ``accuracy'' and ``error'' interchangeably and this can sometimes cause confusion to the reader. For this reason we use the word ``error'' in any case where the meaning could be unclear. All angles are reported in degrees. In this paper, error was determined as the average euclidean distance between the ground truth and predicted angles on the left-out, person-exclusive test set. 

Multiple deep neural networks were compared for eye gaze estimation using deep neural networks. The publicly available MPII Gaze dataset was used for all experiments except for the first where the UT Multiview dataset is also used. 

\subsection{MPII Gaze Dataset Details}
The MPII gaze dataset is a large collection of 213659 images captured under unconstrained conditions from 15 subjects over several days. The images are collected under multiple illumination conditions. Some of the subjects wear spectacles and some do not. The images were captured at various gaze angles, recorded by software running on the participant's laptops. In each session, the subjects were asked to look at random sequences of 20 onscreen positions and to confirm their attentiveness, the subjects were asked to press the space bar once the onscreen target was disappearing. 

The dataset contains eye and head features and target (gaze angle) values for every participant.  To use MPII Gaze, the authors suggest mapping their reported vector to angles using a Rodrigious transformation, and this has been done for all reported experiments.

\section{Experiments and Results}

In this section, multiple experiments are described to provide insight on 4 primary research questions. These are tested on multiple CNN architectures and are discussed in this section. One of the first research goals was to achieve state of the art test error on a network that could perform inference within 3-15 ms on a typical single proprietary low power consumer embedded device. 

The specific research questions are:
\begin{enumerate}
\item How does an architecture that uses both eyes compare to one that uses one eye in terms of accuracy? 
\item How does simulated camera distance impact eye gaze accuracy for the proposed model? 
\item Can augmentation be used to reduce any negative impacts?
\item Can the proposed hardware-friendly architecture perform with sufficient accuracy and speed? 
\end{enumerate}
First, the intra-dataset, person-exclusive experiments from \cite{39} were duplicated. The same procedure to estimate accuracy was used except the altered accuracy layers were modified to eliminate NaNs by replacing undefined values of the arc cosine function with the largest or smallest valid values as appropriate.  
\subsection{Approach 1: Analysis of Eye Flipping}

In \cite{39}, one network is used for both eyes and one of the eyes is flipped so that the gaze angle is roughly correct. An experiment was designed to see if this flipping had an impact on model accuracy. Six experiments were performed using the UT and MPII-Gaze datasets to see if training on both eyes or just one eye impacted accuracy. By doing experiments with combined and non combined datasets, it was also possible to determine if they had similar distributions, and thus, if combining the two would be helpful for future experiments.

\begin{table}[]
\centering
\caption{Results of Approach 1}
\label{exp1}
\begin{tabular}{ll}
Training set           & Error \\
MPII Left eye only     & 5.9 degrees       \\
MPII+UT Left eye only  & 5.6 degrees       \\
MPII Both eyes         & 7.4 degrees      \\
MPII+UT Both eyes      & 6.5 degrees      \\
MPII Right eye only    & 5.3 degrees       \\
MPII+UT Right eye only & 6.1 degrees      
\end{tabular}
\end{table}

As shown in table \ref{exp1}, the method of individually classifying eye images and simply adjusting the right eye and angles as used in \cite{39} is a limiting factor in accuracy for that method. In both datasets, the performance was increased exclusively using left eyes or right eyes. This suggests that simply flipping the eye as suggested by \cite{39} may be a source of error in their model. 

These results also indicated that the distribution of MPII Gaze and UT-Multiview are sufficiently different that combining the two for training gives no or very little improvement. Because of this, it was decided to use only MPII Gaze for the remaining experiments in this section as UT-Multiview had no significant influence on error. 

\subsection{A New Approach: Dual Eye Channels}

Given the problems identified in the previous subsection with flipping one of the eyes, and not wanting to use two different networks for reasons of efficiency, a new approach involving using both the left and right eyes in separate input channels was investigated. Specifically, the left eye and right eye images are passed to the network in channels 0 and 1 respectively, and the gaze and pose information are averaged between the left and right eye images to create a single gaze and pose vector.  Due to the results in the previous section, which indicated that data from UT did not significantly impact the results, only the MPII-Gaze dataset was used. 

This modified, two channel architecture resulted in a significant increase in accuracy, averaging 4.63 degrees of error between the target and predicted values on unseen individuals. A diagram of this network can be seen in Figure 1.

\begin{figure}[h]
\includegraphics[width=5.5in,trim={0 3.2in 0 0in}]{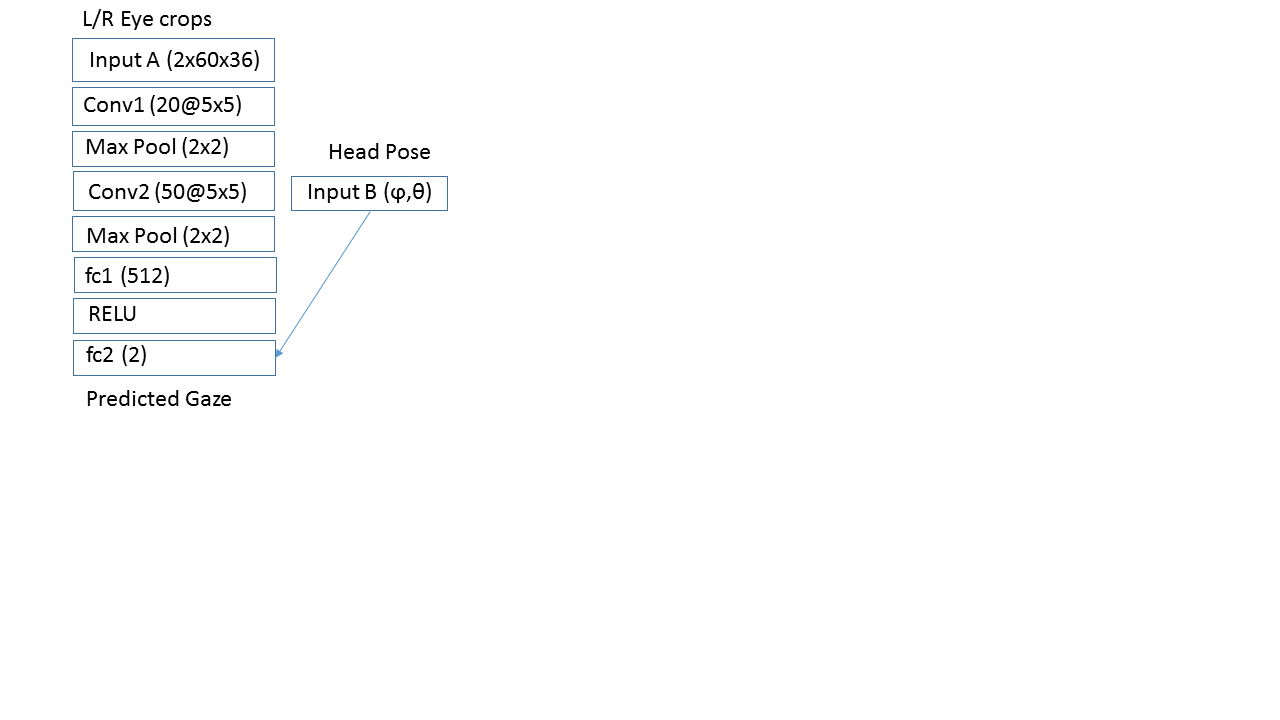}
\caption{Diagram of hardware CNN for initial eye gaze estimation tasks, heavily based on network from \cite{39}}
\end{figure}

\subsection{Can We Reduce the Number of Parameters?}

Deep neural networks can often be made more efficient by reducing the number of parameters but this can sometimes come at the cost of accuracy. To see if reducing the number of parameters was possible without harming accuracy, an experiment was performed to halve the size of all output parameters. This experiment was not allowed to run for the full duration because the exact angle accuracy did not matter, only evidence that the network complexity could be reduced to a point where it would be small enough if necessary. This resulted in an average error of 4.980\% on an unseen individual from the MPII Gaze dataset and indicates that reducing the number of parameters had little impact on accuracy. 

\subsection{Multi Resolution Experiments}

Eye gaze systems in consumer devices must be able to maintain accuracy at a large range of distances. Although MPII Gaze has some variability in distance from the camera, the distances are not realistic for the conditions expected in, for example, a driver monitoring system or a distant cell phone camera. Specifically, it was desired to accommodate realistic distances between the camera and the subject in situations that would be typical in commercial eye trackers that utilize low cost, low resolution cameras. To simulate the loss of information caused by distance, down-sampling was performed on the eye images in MPII-Gaze as follows:

\begin{itemize}
\item Input image 60 x 36 -\textgreater Downscale to 52 x 31 -\textgreater Upscale to 60 x 36 -\textgreater CNN Eye gaze angle 
\item Item Input image 60 x 36 -\textgreater Downscale to 26 x 16 -\textgreater Upscale to 60 x 36 -\textgreater CNN Eye gaze angle.
\end{itemize}

As can be seen in table \ref{disstable}, the network learned a narrow range of distances, and performance deteriorates  when the subject is further from the camera than those in the training set. As a sanity check, an experiment was done to see if the Downscaling algorithm was at fault for the poor results, so in addition to nearest, we also tried bicubic, linear, and LANCZOS from using OpenCV. The experiment showed that the downscaling algorithm used had no influence on the results.

\begin{table}[]
\centering
\caption{Result of distance simulation experiments}
\label{disstable}
\begin{tabular}{ll}
Resolution   & Error \\
60 x 36 & 4.63 degrees  \\
52 x 31      & 9.90 degrees    \\
26 x 16      & 10.10 degrees    
\end{tabular}
\end{table}

This demonstrates that the model is sensitive to changes in distance. In the next section, an experiment is performed to see if data augmentation can be used to improve upon this. 

\subsection{Impact of Random Resizing as Augmentation}

Data augmentation has been shown in many studies to have a large impact on model performance. 

To further improve accuracy, the dataset was augmented with multiple randomly chosen resolutions to match the full range of desired distances. To help reduce the chance that the network would learn the specific interpolation method used, Nearest is used in the training set, but Lanczos filtering is used in the testing set. 

\begin{itemize}
\item Original resolution: 4.918 degrees error
\item 60 x 36 -\textgreater 52 x 31 -\textgreater 60 x 36 : 4.94 degrees error
\item 60 x 36 -\textgreater 26 x 16 -\textgreater 60 x 36: 4.97 degrees error
\end{itemize}

These results indicate that augmenting the images with distances that are likely to be encountered in real world usage situations is an effective way to increase accuracy and succeeds in achieving some invariance to subject distance. 

\subsection{A Quest for Hardware Efficiency and Even Better Accuracy}
It was shown in \cite{simonyan2014very} that two stacked layers of 3x3 convolutions has the same receptive field as a single 5x5 layer, with fewer multiplications. Due to this, one of the requirements was that kernel sizes be 3x3, this required retraining and slight redesign of the network. Several experiments involving architectures with stacked 3x3 kernels were performed using different parameters. The best two architectures were further evaluated, and the best models from each of them were chosen and evaluated on multiple resolutions as shown in table \ref{best}. A diagram of the final architecture used can be seen in figure 2, and a comparison with other published works can be seen in table \ref{Compared}. 
\begin{table}[]
\centering
\caption{Error of proposed model for various resolutions (degrees)}
\label{best}
\begin{tabular}{llp{2cm}lp{2cm}l}
Resolutions & 36 x 60  &  31 x 52  &  26 x 16\\
Best              & 3.650 & 4.1690                                         & 4.240                                         \\
Second best       & 4.100  & 4.324                                         & 4.366                                         \\
Benchmark  & 4.917 & 4.940                                         & 4.970                                        
\end{tabular}
\end{table}
\begin{figure}[h]
\includegraphics[width=5.5in,trim={0 1.1in 0 0in}]{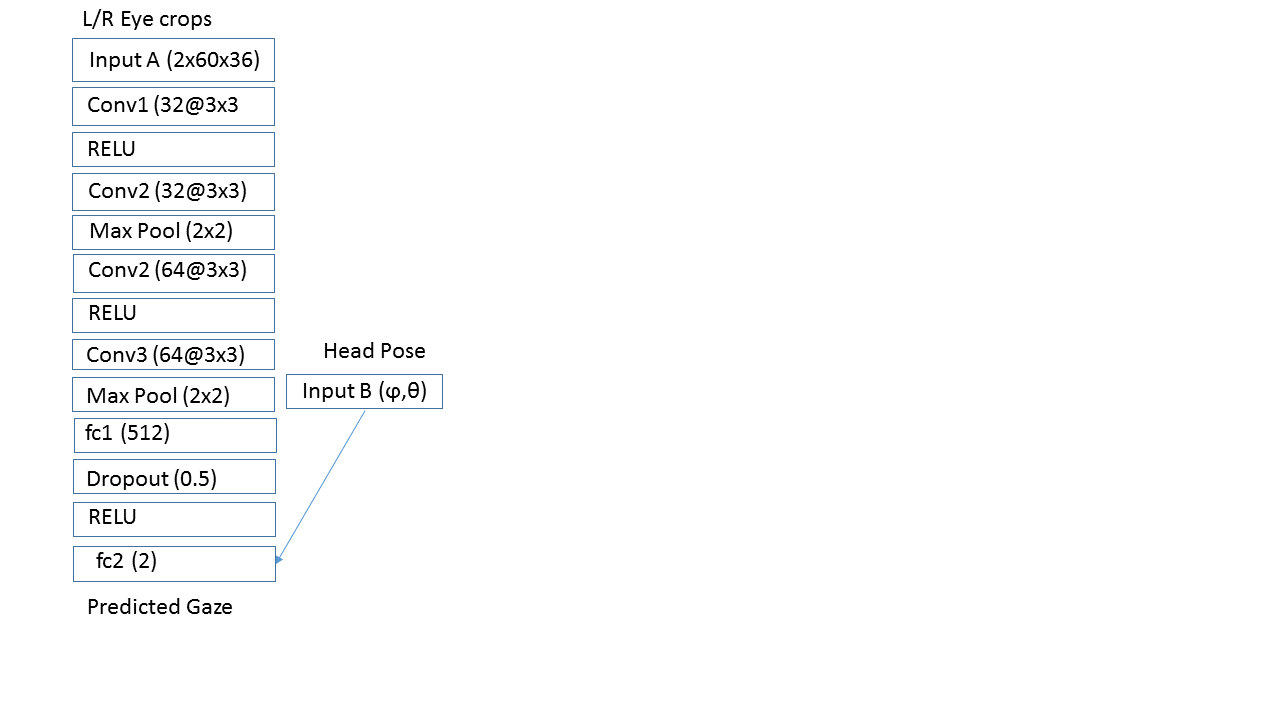}
\caption{Diagram of hardware optimized CNN for eye gaze estimation corresponding to the accuracies in table \ref{best}.}
\end{figure}

\begin{table}[]
\centering
\caption{Comparison of proposed model with other published works on the MPII-Gaze database}
\label{Compared}
\begin{tabular}{|l|l|}
\hline
\textbf{Citation} & \textbf{Error (degrees)} \\ \hline
\cite{52}                & 9.96                     \\ \hline
\cite{50}                & 9.58                     \\ \hline
\cite{53}                & 7.8                      \\ \hline
\cite{49}                & 7.1                      \\ \hline
\cite{39}                & 6                        \\ \hline
\cite{41}                & 4.8                      \\ \hline
Proposed          & 3.64                     \\ \hline
\end{tabular}
\end{table}

\section{Conclusion}

Our results show that using information from both eyes in the neural network can increase accuracy. This is demonstrated in section V, where adding additional eye information from the opposite eye enabled improved results over individual eyes, helping the network make sense of low quality images with ambiguous gaze. As expected, in all cases, the deeper network had the best performance. This research demonstrated the sensitivity of such models to variations in distance and how data augmentation can be used to overcome this. Most importantly, a new compact hardware-friendly architecture designed for use in small consumer electronics has been introduced and evaluated on the eye gaze task. 

When evaluated on MPII Gaze, the proposed model performs favorably (see table \ref{Compared} even when compared with much larger networks in the literature.

Since augmentation resulted in a significant improvement in accuracy, it may be fruitful to try other types of augmentation such as Generative Adversarial Networks (GANS) with landmarks, \cite{bazrafkan2018face} and Smart Augmentation (SA) \cite{DBLP:journals/access/LemleyBC17} in future work. This will either require modifying such methods to work on regression problems or translating the eye gaze problem into a classification task for the purpose of generating augmented data\cite{lemley2018learning} and then back to a regression task. Additionally, there are plans to investigate whether temporal information\cite{lemley2017transfer} can be used to further increase the accuracy without sacrificing the need for performance as it has been shown to increase performance in DMS systems. 


%
\section*{Acknowledgment}

This research is funded under the SFI Strategic Partnership Program by Science Foundation Ireland (SFI) and FotoNation Ltd. Project ID: 13/SPP/I2868 on Next Generation Imaging for Smartphone and Embedded Platforms. 
This work is also supported by an Irish Research Council Employment Based Programme Award. Project ID: EBPPG/2016/280.

\ifCLASSOPTIONcaptionsoff
  \newpage
\fi



%


\bibliography{refs} 

\begin{thebibliography}{10}
\providecommand{\url}[1]{#1}
\csname url@samestyle\endcsname
\providecommand{\newblock}{\relax}
\providecommand{\bibinfo}[2]{#2}
\providecommand{\BIBentrySTDinterwordspacing}{\spaceskip=0pt\relax}
\providecommand{\BIBentryALTinterwordstretchfactor}{4}
\providecommand{\BIBentryALTinterwordspacing}{\spaceskip=\fontdimen2\font plus
\BIBentryALTinterwordstretchfactor\fontdimen3\font minus
  \fontdimen4\font\relax}
\providecommand{\BIBforeignlanguage}[2]{{%
\expandafter\ifx\csname l@#1\endcsname\relax
\typeout{** WARNING: IEEEtran.bst: No hyphenation pattern has been}%
\typeout{** loaded for the language `#1'. Using the pattern for}%
\typeout{** the default language instead.}%
\else
\language=\csname l@#1\endcsname
\fi
#2}}
\providecommand{\BIBdecl}{\relax}
\BIBdecl

\bibitem{1}
Y.~Liang, M.~L. Reyes, and J.~D. Lee, ``Real-time detection of driver cognitive
  distraction using support vector machines,'' \emph{IEEE Transactions on
  Intelligent Transportation Systems}, vol.~8, no.~2, pp. 340--350, June 2007.

\bibitem{2}
\BIBentryALTinterwordspacing
E.~Wood and A.~Bulling, ``Eyetab: Model-based gaze estimation on unmodified
  tablet computers,'' in \emph{Proceedings of the Symposium on Eye Tracking
  Research and Applications}, ser. ETRA '14.\hskip 1em plus 0.5em minus
  0.4em\relax New York, NY, USA: ACM, 2014, pp. 207--210. [Online]. Available:
  \url{http://doi.acm.org/10.1145/2578153.2578185}
\BIBentrySTDinterwordspacing

\bibitem{3}
A.~Tawari and M.~M. Trivedi, ``Robust and continuous estimation of driver gaze
  zone by dynamic analysis of multiple face videos,'' in \emph{2014 IEEE
  Intelligent Vehicles Symposium Proceedings}, June 2014, pp. 344--349.

\bibitem{4}
\BIBentryALTinterwordspacing
V.~Vaitukaitis and A.~Bulling, ``Eye gesture recognition on portable devices,''
  in \emph{Proceedings of the 2012 ACM Conference on Ubiquitous Computing},
  ser. UbiComp '12.\hskip 1em plus 0.5em minus 0.4em\relax New York, NY, USA:
  ACM, 2012, pp. 711--714. [Online]. Available:
  \url{http://doi.acm.org/10.1145/2370216.2370370}
\BIBentrySTDinterwordspacing

\bibitem{5}
S.-W. Shih and J.~Liu, ``A novel approach to 3-d gaze tracking using stereo
  cameras,'' \emph{IEEE Transactions on Systems, Man, and Cybernetics, Part B
  (Cybernetics)}, vol.~34, no.~1, pp. 234--245, Feb 2004.

\bibitem{lecun1989backpropagation}
Y.~LeCun, B.~Boser, J.~S. Denker, D.~Henderson, R.~E. Howard, W.~Hubbard, and
  L.~D. Jackel, ``Backpropagation applied to handwritten zip code
  recognition,'' \emph{Neural computation}, vol.~1, no.~4, pp. 541--551, 1989.

\bibitem{lemley2017deep}
J.~Lemley, S.~Bazrafkan, and P.~Corcoran, ``Deep learning for consumer devices
  and services: Pushing the limits for machine learning, artificial
  intelligence, and computer vision.'' \emph{IEEE Consumer Electronics
  Magazine}, vol.~6, no.~2, pp. 48--56, 2017.

\bibitem{hansen2010eye}
D.~W. Hansen and Q.~Ji, ``In the eye of the beholder: A survey of models for
  eyes and gaze,'' \emph{IEEE transactions on pattern analysis and machine
  intelligence}, vol.~32, no.~3, pp. 478--500, 2010.

\bibitem{Kar2017}
A.~Kar and P.~Corcoran, ``A review and analysis of eye-gaze estimation systems,
  algorithms and performance evaluation methods in consumer platforms,''
  \emph{IEEE Access}, vol.~5, pp. 16\,495--16\,519, 2017.

\bibitem{coutinho2012augmenting}
F.~L. Coutinho and C.~H. Morimoto, ``Augmenting the robustness of cross-ratio
  gaze tracking methods to head movement,'' in \emph{Proceedings of the
  Symposium on Eye Tracking Research and Applications}.\hskip 1em plus 0.5em
  minus 0.4em\relax ACM, 2012, pp. 59--66.

\bibitem{chen2015eye}
S.~Chen and C.~Liu, ``Eye detection using discriminatory haar features and a
  new efficient svm,'' \emph{Image and Vision Computing}, vol.~33, pp. 68--77,
  2015.

\bibitem{22}
H.~chuan Lu, C.~Wang, and Y.~w.~Chen, ``Gaze tracking by binocular vision and
  lbp features,'' in \emph{2008 19th International Conference on Pattern
  Recognition}, Dec 2008, pp. 1--4.

\bibitem{6}
\BIBentryALTinterwordspacing
P.~Blignaut, ``Mapping the pupil-glint vector to gaze coordinates in a simple
  video-based eye tracker,'' \emph{Journal of Eye Movement Research}, vol.~7,
  no.~1, 2013. [Online]. Available:
  \url{https://bop.unibe.ch/index.php/JEMR/article/view/2373}
\BIBentrySTDinterwordspacing

\bibitem{7}
\BIBentryALTinterwordspacing
Z.~Zhu and Q.~Ji, ``Eye and gaze tracking for interactive graphic display,''
  \emph{Machine Vision and Applications}, vol.~15, no.~3, pp. 139--148, Jul
  2004. [Online]. Available: \url{https://doi.org/10.1007/s00138-004-0139-4}
\BIBentrySTDinterwordspacing

\bibitem{8}
C.~Jian-nan, Z.~Chuang, Y.~Yan-tao, L.~Yang, and Z.~Han, ``Eye gaze calculation
  based on nonlinear polynomial and generalized regression neural network,'' in
  \emph{2009 Fifth International Conference on Natural Computation}, vol.~3,
  Aug 2009, pp. 617--623.

\bibitem{11}
\BIBentryALTinterwordspacing
J.~Wang, G.~Zhang, and J.~Shi, ``2d gaze estimation based on pupil-glint vector
  using an artificial neural network,'' \emph{Applied Sciences}, vol.~6, no.~6,
  2016. [Online]. Available: \url{http://www.mdpi.com/2076-3417/6/6/174}
\BIBentrySTDinterwordspacing

\bibitem{9}
\BIBentryALTinterwordspacing
Chunfei, K.-A. Ma, B.-D. Choi, and S.-J.~K. Choi, ``Robust remote gaze
  estimation method based on multiple geometric transforms,'' \emph{Optical
  Engineering}, vol.~54, pp. 54 -- 54 -- 7, 2015. [Online]. Available:
  \url{https://doi.org/10.1117/1.OE.54.8.083103}
\BIBentrySTDinterwordspacing

\bibitem{10}
Z.~Zhu, Q.~Ji, and K.~P. Bennett, ``Nonlinear eye gaze mapping function
  estimation via support vector regression,'' in \emph{18th International
  Conference on Pattern Recognition (ICPR'06)}, vol.~1, 2006, pp. 1132--1135.

\bibitem{12}
J.~Zhu and J.~Yang, ``Subpixel eye gaze tracking,'' in \emph{Proceedings of
  Fifth IEEE International Conference on Automatic Face Gesture Recognition},
  May 2002, pp. 124--129.

\bibitem{13}
E.~D. Guestrin and M.~Eizenman, ``General theory of remote gaze estimation
  using the pupil center and corneal reflections,'' \emph{IEEE Transactions on
  Biomedical Engineering}, vol.~53, no.~6, pp. 1124--1133, June 2006.

\bibitem{14}
\BIBentryALTinterwordspacing
T.~Ohno and N.~Mukawa, ``A free-head, simple calibration, gaze tracking system
  that enables gaze-based interaction,'' in \emph{Proceedings of the 2004
  Symposium on Eye Tracking Research \& Applications}, ser. ETRA '04.\hskip 1em
  plus 0.5em minus 0.4em\relax New York, NY, USA: ACM, 2004, pp. 115--122.
  [Online]. Available: \url{http://doi.acm.org/10.1145/968363.968387}
\BIBentrySTDinterwordspacing

\bibitem{15}
Z.~Zhu and Q.~Ji, ``Novel eye gaze tracking techniques under natural head
  movement,'' \emph{IEEE Transactions on Biomedical Engineering}, vol.~54,
  no.~12, pp. 2246--2260, Dec 2007.

\bibitem{16}
D.~Beymer and M.~Flickner, ``Eye gaze tracking using an active stereo head,''
  in \emph{2003 IEEE Computer Society Conference on Computer Vision and Pattern
  Recognition, 2003. Proceedings.}, vol.~2, June 2003, pp. II--451--8 vol.2.

\bibitem{17}
\BIBentryALTinterwordspacing
T.~Nagamatsu, J.~Kamahara, and N.~Tanaka, ``Calibration-free gaze tracking
  using a binocular 3d eye model,'' in \emph{CHI '09 Extended Abstracts on
  Human Factors in Computing Systems}, ser. CHI EA '09.\hskip 1em plus 0.5em
  minus 0.4em\relax New York, NY, USA: ACM, 2009, pp. 3613--3618. [Online].
  Available: \url{http://doi.acm.org/10.1145/1520340.1520543}
\BIBentrySTDinterwordspacing

\bibitem{18}
D.~Model and M.~Eizenman, ``User-calibration-free remote eye-gaze tracking
  system with extended tracking range,'' in \emph{2011 24th Canadian Conference
  on Electrical and Computer Engineering(CCECE)}, May 2011, pp.
  001\,268--001\,271.

\bibitem{19}
K.~Wang and Q.~Ji, ``Real time eye gaze tracking with kinect,'' in \emph{2016
  23rd International Conference on Pattern Recognition (ICPR)}, Dec 2016, pp.
  2752--2757.

\bibitem{20}
X.~Zhou, H.~Cai, Z.~Shao, H.~Yu, and H.~Liu, ``3d eye model-based gaze
  estimation from a depth sensor,'' in \emph{2016 IEEE International Conference
  on Robotics and Biomimetics (ROBIO)}, Dec 2016, pp. 369--374.

\bibitem{21}
\BIBentryALTinterwordspacing
Y.-L. Wu, C.-T. Yeh, W.-C. Hung, and C.-Y. Tang, ``Gaze direction estimation
  using support vector machine with active appearance model,'' \emph{Multimedia
  Tools and Applications}, vol.~70, no.~3, pp. 2037--2062, Jun 2014. [Online].
  Available: \url{https://doi.org/10.1007/s11042-012-1220-z}
\BIBentrySTDinterwordspacing

\bibitem{23}
C.~M. Yilmaz and C.~Kose, ``Local binary pattern histogram features for
  on-screen eye-gaze direction estimation and a comparison of appearance based
  methods,'' in \emph{2016 39th International Conference on Telecommunications
  and Signal Processing (TSP)}, June 2016, pp. 693--696.

\bibitem{24}
\BIBentryALTinterwordspacing
S.~Chen and C.~Liu, ``Eye detection using discriminatory haar features and a
  new efficient svm,'' \emph{Image Vision Comput.}, vol.~33, no.~C, pp. 68--77,
  Jan. 2015. [Online]. Available:
  \url{http://dx.doi.org/10.1016/j.imavis.2014.10.007}
\BIBentrySTDinterwordspacing

\bibitem{25}
Y.~Li, X.~Xu, N.~Mu, and L.~Chen, ``Eye-gaze tracking system by haar cascade
  classifier,'' in \emph{2016 IEEE 11th Conference on Industrial Electronics
  and Applications (ICIEA)}, June 2016, pp. 564--567.

\bibitem{27}
\BIBentryALTinterwordspacing
T.~Schneider, B.~Schauerte, and R.~Stiefelhagen, ``Manifold alignment for
  person independent appearance-based gaze estimation,'' in \emph{Proceedings
  of the 2014 22Nd International Conference on Pattern Recognition}, ser. ICPR
  '14.\hskip 1em plus 0.5em minus 0.4em\relax Washington, DC, USA: IEEE
  Computer Society, 2014, pp. 1167--1172. [Online]. Available:
  \url{http://dx.doi.org/10.1109/ICPR.2014.210}
\BIBentrySTDinterwordspacing

\bibitem{26}
S.~Baluja and D.~Pomerleau, ``Non-intrusive gaze tracking using artificial
  neural networks,'' in \emph{Advances in Neural Information Processing
  Systems}, 1994, pp. 753--760.

\bibitem{28}
Y.~Fu, W.~P. Zhu, and D.~Massicotte, ``A gaze tracking scheme with low
  resolution image,'' in \emph{2013 IEEE 11th International New Circuits and
  Systems Conference (NEWCAS)}, June 2013, pp. 1--4.

\bibitem{29}
A.~George and A.~Routray, ``Fast and accurate algorithm for eye localisation
  for gaze tracking in low-resolution images,'' \emph{IET Computer Vision},
  vol.~10, no.~7, pp. 660--669, 2016.

\bibitem{30}
D.~Gonzalez-Jimenez and J.~L. Alba-Castro, ``Shape-driven gabor jets for face
  description and authentication,'' \emph{IEEE Transactions on Information
  Forensics and Security}, vol.~2, no.~4, pp. 769--780, Dec 2007.

\bibitem{31}
\BIBentryALTinterwordspacing
M.~Tonsen, J.~Steil, Y.~Sugano, and A.~Bulling, ``Invisibleeye: Mobile eye
  tracking using multiple low-resolution cameras and learning-based gaze
  estimation,'' \emph{Proc. ACM Interact. Mob. Wearable Ubiquitous Technol.},
  vol.~1, no.~3, pp. 106:1--106:21, Sep. 2017. [Online]. Available:
  \url{http://doi.acm.org/10.1145/3130971}
\BIBentrySTDinterwordspacing

\bibitem{32}
\BIBentryALTinterwordspacing
Y.-T. Lin, R.-Y. Lin, Y.-C. Lin, and G.~C. Lee, ``Real-time eye-gaze estimation
  using a low-resolution webcam,'' \emph{Multimedia Tools Appl.}, vol.~65,
  no.~3, pp. 543--568, Aug. 2013. [Online]. Available:
  \url{http://dx.doi.org/10.1007/s11042-012-1202-1}
\BIBentrySTDinterwordspacing

\bibitem{33}
Y.~Ono, T.~Okabe, and Y.~Sato, ``Gaze estimation from low resolution images,''
  in \emph{Advances in Image and Video Technology}, L.-W. Chang and W.-N. Lie,
  Eds.\hskip 1em plus 0.5em minus 0.4em\relax Berlin, Heidelberg: Springer
  Berlin Heidelberg, 2006, pp. 178--188.

\bibitem{36}
A.~George and A.~Routray, ``Real-time eye gaze direction classification using
  convolutional neural network,'' in \emph{2016 International Conference on
  Signal Processing and Communications (SPCOM)}, June 2016, pp. 1--5.

\bibitem{37}
R.~Konrad, S.~Shrestha, and P.~Varma, ``Near-eye display gaze tracking via
  convolutional neural networks.''

\bibitem{38}
\BIBentryALTinterwordspacing
Y.~Wang, T.~Shen, G.~Yuan, J.~Bian, and X.~Fu, ``Appearance-based gaze
  estimation using deep features and random forest regression,''
  \emph{Know.-Based Syst.}, vol. 110, no.~C, pp. 293--301, Oct. 2016. [Online].
  Available: \url{https://doi.org/10.1016/j.knosys.2016.07.038}
\BIBentrySTDinterwordspacing

\bibitem{41}
X.~Zhang, Y.~Sugano, M.~Fritz, and A.~Bulling, ``It's written all over your
  face: Full-face appearance-based gaze estimation,'' in \emph{2017 IEEE
  Conference on Computer Vision and Pattern Recognition Workshops (CVPRW)},
  July 2017, pp. 2299--2308.

\bibitem{42}
H.~Deng and W.~Zhu, ``Monocular free-head 3d gaze tracking with deep learning
  and geometry constraints,'' in \emph{2017 IEEE International Conference on
  Computer Vision (ICCV)}, Oct 2017, pp. 3162--3171.

\bibitem{39}
X.~Zhang, Y.~Sugano, M.~Fritz, and A.~Bulling, ``Appearance-based gaze
  estimation in the wild,'' in \emph{2015 IEEE Conference on Computer Vision
  and Pattern Recognition (CVPR)}, June 2015, pp. 4511--4520.

\bibitem{43}
K.~Krafka, A.~Khosla, P.~Kellnhofer, H.~Kannan, S.~Bhandarkar, W.~Matusik, and
  A.~Torralba, ``Eye tracking for everyone,'' in \emph{2016 IEEE Conference on
  Computer Vision and Pattern Recognition (CVPR)}, June 2016, pp. 2176--2184.

\bibitem{44}
H.~Park and D.~Kim, ``Gaze classification on a mobile device by using deep
  belief networks,'' in \emph{2015 3rd IAPR Asian Conference on Pattern
  Recognition (ACPR)}, Nov 2015, pp. 685--689.

\bibitem{45}
\BIBentryALTinterwordspacing
C.~Zhang, R.~Yao, and J.~Cai, ``Efficient eye typing with 9-direction gaze
  estimation,'' \emph{Multimedia Tools and Applications}, Nov 2017. [Online].
  Available: \url{https://doi.org/10.1007/s11042-017-5426-y}
\BIBentrySTDinterwordspacing

\bibitem{46}
X.~Wu, J.~Li, Q.~Wu, and J.~Sun, ``Appearance-based gaze block estimation via
  cnn classification,'' in \emph{2017 IEEE 19th International Workshop on
  Multimedia Signal Processing (MMSP)}, Oct 2017, pp. 1--5.

\bibitem{56}
\BIBentryALTinterwordspacing
Y.~Sugano, Y.~Matsushita, and Y.~Sato, ``Learning-by-synthesis for
  appearance-based 3d gaze estimation,'' in \emph{Proceedings of the 2014 IEEE
  Conference on Computer Vision and Pattern Recognition}, ser. CVPR '14.\hskip
  1em plus 0.5em minus 0.4em\relax Washington, DC, USA: IEEE Computer Society,
  2014, pp. 1821--1828. [Online]. Available:
  \url{http://dx.doi.org/10.1109/CVPR.2014.235}
\BIBentrySTDinterwordspacing

\bibitem{57}
\BIBentryALTinterwordspacing
K.~A. Funes~Mora, F.~Monay, and J.-M. Odobez, ``Eyediap: A database for the
  development and evaluation of gaze estimation algorithms from rgb and rgb-d
  cameras,'' in \emph{Proceedings of the Symposium on Eye Tracking Research and
  Applications}, ser. ETRA '14.\hskip 1em plus 0.5em minus 0.4em\relax New
  York, NY, USA: ACM, 2014, pp. 255--258. [Online]. Available:
  \url{http://doi.acm.org/10.1145/2578153.2578190}
\BIBentrySTDinterwordspacing

\bibitem{59}
\BIBentryALTinterwordspacing
Q.~Huang, A.~Veeraraghavan, and A.~Sabharwal, ``Tabletgaze: dataset and
  analysis for unconstrained appearance-based gaze estimation in mobile
  tablets,'' \emph{Machine Vision and Applications}, vol.~28, no.~5, pp.
  445--461, Aug 2017. [Online]. Available:
  \url{https://doi.org/10.1007/s00138-017-0852-4}
\BIBentrySTDinterwordspacing

\bibitem{61}
U.~Weidenbacher, G.~Layher, P.~M. Strauss, and H.~Neumann, ``A comprehensive
  head pose and gaze database,'' in \emph{2007 3rd IET International Conference
  on Intelligent Environments}, Sept 2007, pp. 455--458.

\bibitem{62}
\BIBentryALTinterwordspacing
C.~D. McMurrough, V.~Metsis, J.~Rich, and F.~Makedon, ``An eye tracking dataset
  for point of gaze detection,'' in \emph{Proceedings of the Symposium on Eye
  Tracking Research and Applications}, ser. ETRA '12.\hskip 1em plus 0.5em
  minus 0.4em\relax New York, NY, USA: ACM, 2012, pp. 305--308. [Online].
  Available: \url{http://doi.acm.org/10.1145/2168556.2168622}
\BIBentrySTDinterwordspacing

\bibitem{63}
Q.~He, X.~Hong, X.~Chai, J.~Holappa, G.~Zhao, X.~Chen, and M.~Pietik{\"a}inen,
  ``Omeg: Oulu multi-pose eye gaze dataset,'' in \emph{Image Analysis}, R.~R.
  Paulsen and K.~S. Pedersen, Eds.\hskip 1em plus 0.5em minus 0.4em\relax Cham:
  Springer International Publishing, 2015, pp. 418--427.

\bibitem{65}
\BIBentryALTinterwordspacing
S.~Asteriadis, D.~Soufleros, K.~Karpouzis, and S.~Kollias, ``A natural head
  pose and eye gaze dataset,'' in \emph{Proceedings of the International
  Workshop on Affective-Aware Virtual Agents and Social Robots}, ser. AFFINE
  '09.\hskip 1em plus 0.5em minus 0.4em\relax New York, NY, USA: ACM, 2009, pp.
  1:1--1:4. [Online]. Available:
  \url{http://doi.acm.org/10.1145/1655260.1655261}
\BIBentrySTDinterwordspacing

\bibitem{48}
S.~D. Iyer and H.~Ramasangu, ``Hybrid lasso and neural network estimator for
  gaze estimation,'' in \emph{2016 IEEE Region 10 Conference (TENCON)}, Nov
  2016, pp. 2579--2582.

\bibitem{49}
S.~Nie, M.~Zheng, and Q.~Ji, ``The deep regression bayesian network and its
  applications: Probabilistic deep learning for computer vision,'' \emph{IEEE
  Signal Processing Magazine}, vol.~35, no.~1, pp. 101--111, Jan 2018.

\bibitem{50}
\BIBentryALTinterwordspacing
E.~Wood, T.~Baltru\v{s}aitis, L.-P. Morency, P.~Robinson, and A.~Bulling,
  ``Learning an appearance-based gaze estimator from one million synthesised
  images,'' in \emph{Proceedings of the Ninth Biennial ACM Symposium on Eye
  Tracking Research \& Applications}, ser. ETRA '16.\hskip 1em plus 0.5em minus
  0.4em\relax New York, NY, USA: ACM, 2016, pp. 131--138. [Online]. Available:
  \url{http://doi.acm.org/10.1145/2857491.2857492}
\BIBentrySTDinterwordspacing

\bibitem{51}
\BIBentryALTinterwordspacing
E.~Wood, T.~Baltruaitis, X.~Zhang, Y.~Sugano, P.~Robinson, and A.~Bulling,
  ``Rendering of eyes for eye-shape registration and gaze estimation,'' in
  \emph{Proceedings of the 2015 IEEE International Conference on Computer
  Vision (ICCV)}, ser. ICCV '15.\hskip 1em plus 0.5em minus 0.4em\relax
  Washington, DC, USA: IEEE Computer Society, 2015, pp. 3756--3764. [Online].
  Available: \url{http://dx.doi.org/10.1109/ICCV.2015.428}
\BIBentrySTDinterwordspacing

\bibitem{52}
T.~Baltrusaitis, P.~Robinson, and L.~P. Morency, ``Openface: An open source
  facial behavior analysis toolkit,'' in \emph{2016 IEEE Winter Conference on
  Applications of Computer Vision (WACV)}, March 2016, pp. 1--10.

\bibitem{53}
A.~Shrivastava, T.~Pfister, O.~Tuzel, J.~Susskind, W.~Wang, and R.~Webb,
  ``Learning from simulated and unsupervised images through adversarial
  training,'' in \emph{2017 IEEE Conference on Computer Vision and Pattern
  Recognition (CVPR)}, July 2017, pp. 2242--2251.

\bibitem{54}
\BIBentryALTinterwordspacing
X.~Zhang, Y.~Sugano, and A.~Bulling, ``Everyday eye contact detection using
  unsupervised gaze target discovery,'' in \emph{Proceedings of the 30th Annual
  ACM Symposium on User Interface Software and Technology}, ser. UIST
  2017.\hskip 1em plus 0.5em minus 0.4em\relax New York, NY, USA: ACM, 2017,
  pp. 193--203. [Online]. Available:
  \url{http://doi.acm.org/10.1145/3126594.3126614}
\BIBentrySTDinterwordspacing

\bibitem{55}
S.~J. Baek, K.~A. Choi, C.~Ma, Y.~H. Kim, and S.~J. Ko, ``Eyeball model-based
  iris center localization for visible image-based eye-gaze tracking systems,''
  \emph{IEEE Transactions on Consumer Electronics}, vol.~59, no.~2, pp.
  415--421, May 2013.

\bibitem{simonyan2014very}
K.~Simonyan and A.~Zisserman, ``Very deep convolutional networks for
  large-scale image recognition,'' \emph{arXiv preprint arXiv:1409.1556}, 2014.

\bibitem{bazrafkan2018face}
S.~Bazrafkan, H.~Javidnia, and P.~Corcoran, ``Face synthesis with landmark
  points from generative adversarial networks and inverse latent space
  mapping,'' \emph{arXiv preprint arXiv:1802.00390}, 2018.

\bibitem{DBLP:journals/access/LemleyBC17}
\BIBentryALTinterwordspacing
J.~Lemley, S.~Bazrafkan, and P.~Corcoran, ``Smart augmentation learning an
  optimal data augmentation strategy,'' \emph{{IEEE} Access}, vol.~5, pp.
  5858--5869, 2017. [Online]. Available:
  \url{https://doi.org/10.1109/ACCESS.2017.2696121}
\BIBentrySTDinterwordspacing

\bibitem{lemley2018learning}
------, ``Learning data augmentation for consumer devices and services,'' in
  \emph{Consumer Electronics (ICCE), 2018 IEEE International Conference
  on}.\hskip 1em plus 0.5em minus 0.4em\relax IEEE, 2018, pp. 1--3.

\bibitem{lemley2017transfer}
------, ``Transfer learning of temporal information for driver action
  classification,'' in \emph{The 28th Modern Artificial Intelligence and
  Cognitive Science Conference (MAICS)}, 2017.

\end{thebibliography}
\bibliographystyle{IEEEtran}

%
%

%

\begin{IEEEbiography}[{\includegraphics[width=1in,height=1.25in,clip,keepaspectratio]{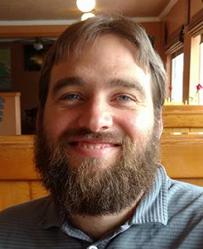}}]{Joseph Lemley}
Joseph Lemley received a B.S. degree in computer science and the Masters degree in computational science from Central Washington University in 2006 and 2016, respectively. He is currently pursuing a Ph.D. with the National University of Ireland Galway. His field of work is machine learning using deep neural networks for tasks related to computer vision. His Ph.D. is funded by FotoNation, Ltd., under the IRCSET Employment Ph.D. Program.
\end{IEEEbiography}

\begin{IEEEbiography}[{\includegraphics[width=1in,height=1.25in,clip,keepaspectratio]{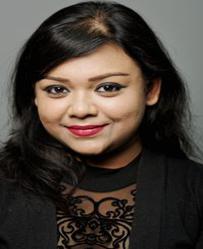}}]{Anurada Kar}
Anuradha Kar is currently pursuing the Ph.D. degree with the National University of Ireland Galway and the Center for Cognitive, Connected, and Computational Imaging. Her research interests include human computer interaction and computational imaging. She is involved in eye gaze tracking-addressing the issues of accuracy and performance evaluation of gaze estimation systems in various platforms.
\end{IEEEbiography}

\begin{IEEEbiography}[{\includegraphics[width=1in,height=1.25in,clip,keepaspectratio]{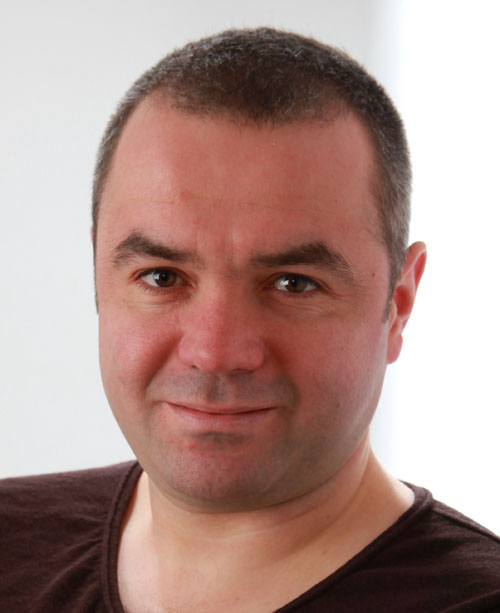}}]{Alexandru Drimbarean}
Alexandru Drimbarean is the Vice President of Advanced Research team at FotoNation Ireland focusing on developing innovative computer vision and machine learning technologies for mobile, biometrics and automotive applications. Alexandru received his B.S in Electronic Engineering in Brasov Romania followed by an M.Sc. in Electronic Science at N.U.I Galway in 2002. His interests include image processing and understanding as well computer vision and machine learning. Alexandru has authored several journal articles as well as more than 30 patents.
\end{IEEEbiography}

\begin{IEEEbiography}[{\includegraphics[width=1in,height=1.25in,clip,keepaspectratio]{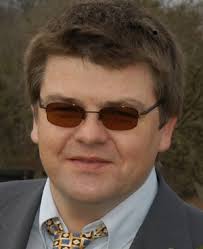}}]{Peter Corcoran}
Peter Corcoran is a Fellow of IEEE, past Editor-in-Chief of IEEE Consumer Electronics Magazine and a Professor at NUI Galway. His research interests include biometrics, imaging, deep learning, edge-AI and consumer electronics. He is co-author on 350+ technical publications and co-inventor on more than 300 granted US patents. In addition to his academic career, he is an occasional entrepreneur, industry consultant and compulsive inventor.
\end{IEEEbiography}





\end{document}